\documentclass[review]{elsarticle}

\usepackage{amsmath}
\usepackage{mathrsfs}
\usepackage{amsfonts,amssymb}
\usepackage{amsthm}
\usepackage{array}
\usepackage{url}
\usepackage{mathtools}
\usepackage{graphicx}
\usepackage{color}
\usepackage{float}
\usepackage{multirow}  
\usepackage{CJK}   
\usepackage{indentfirst}
\usepackage{subfigure}   
\usepackage{ulem}
\usepackage{caption}
\usepackage{hyperref}
\usepackage{booktabs} 

\journal{Journal of \LaTeX\ Templates}

\begin{document}

\begin{frontmatter}

\title{MFIF-GAN: A New Generative Adversarial Network for Multi-Focus Image Fusion}

\author{Yicheng Wang}
\author{Shuang Xu}
\author[]{Junmin Liu\corref{mycorrespondingauthor}}
\cortext[mycorrespondingauthor]{Corresponding author}
\ead{junminliu@mail.xjtu.edu.cn}
\author{Zixiang Zhao}
\author{Chunxia Zhang}
\author{Jiangshe Zhang}
\address{School of Mathematics and Statistics, Xi'an Jiaotong University}

\begin{abstract}
    Multi-Focus Image Fusion (MFIF) is a promising image enhancement technique to 
    obtain all-in-focus images meeting visual needs and it is a precondition for
    other computer vision tasks.
    One of the research trends of MFIF is to avoid the defocus spread effect
    (DSE) around the focus/defocus boundary (FDB).
    In this paper, we propose a network termed MFIF-GAN to attenuate the DSE by 
    generating focus maps in which the foreground region are correctly larger than 
    the corresponding objects.
    The \textit{Squeeze} and \textit{Excitation} Residual module is employed in the network.
    By combining the prior knowledge of training condition, 
    this network is trained on a synthetic dataset
    based on an $\alpha$-matte model. In addition, the reconstruction 
    and gradient regularization terms are combined in the loss functions to 
    enhance the boundary details and improve the quality of fused images. 
    Extensive experiments demonstrate that the MFIF-GAN outperforms several 
    state-of-the-art (SOTA) methods in visual perception, quantitative analysis 
    as well as efficiency.
    Moreover, the edge diffusion and contraction module is firstly proposed 
    to verify that focus maps generated by our method are accurate at the pixel level.
\end{abstract}


\end{frontmatter}


\section{Introduction}

\begin{figure}[htbp]
    \centering
    \subfigure[]{
    \includegraphics[scale = 0.15]{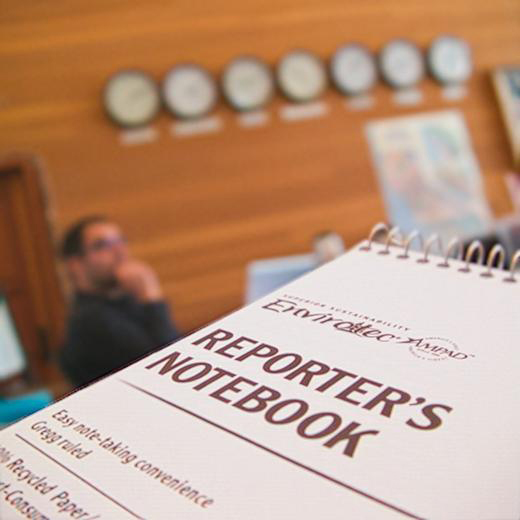}\label{image_a}}
    \subfigure[]{
    \includegraphics[scale = 0.15]{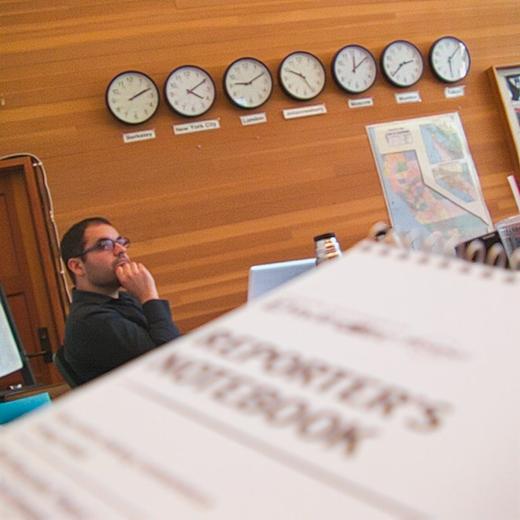}\label{image_b}}
    \subfigure[]{
    \includegraphics[scale = 0.15]{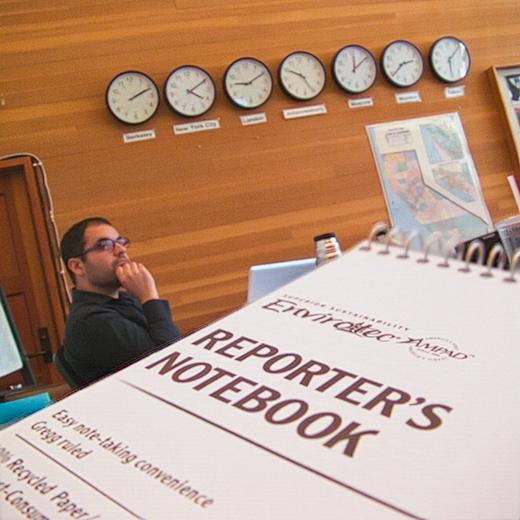}\label{desired_result}}
    \caption{\footnotesize The source of MFIF images and desired fusion result}
\end{figure}

In the field of digital photography, the limited depth-of-field (DOF) leads to
multiple images focused at different regions in the same scene 
and the defocus spread effect (DSE) \cite{MMF_NET}.
As an image enhancement technique, Multi-Focus Image Fusion (MFIF)
has been studied to fuse multi-focus images, as shown in Fig.\ref{image_a} and
\ref{image_b},
so that the fusion result shown in Fig.\ref{desired_result} 
retains the clear information of the sources.
It is a pre-condition for various kinds of computer vision (CV) tasks,
such as localization, object detection, recognition and segmentation
\cite{Image_fusion_SOTA, DOF_morphology}.

The past few decades have witnessed the rapid development of abundant MFIF algorithms.
Generally, the classic MFIF algorithms can be categorized into 
two groups: transform domain and spatial domain methods\cite{Image_fusion}.
The idea of the former is to transform 
the images from the original image space into an abstract feature space 
so that the active 
level of source images can be detected and measured easily. Then a desired image 
is reconstructed from the feature space into the image space after merging the active 
feature according to a certain fusion strategy\cite{MFFW}.
The typical transform domain methods include 
the non-subsampled contourlet transform (NSCT)
\cite{NSCT, NSCT_theory}
, the sparse representation (SR)
\cite{SR, MTSR_SC}
and the combined NSCT-SR
\cite{NSCT_SR}.
The drawback is that these 
algorithms often produce unrealistic 
results, even in the areas far away from the focus/defocus boundary (FDB)
\cite{MMF_NET}.

Based on the assumption that each pixel, block or region is 
either focused or defocused \cite{fusegan},
the spatial domain methods can be classified into three categories: pixel-based, 
block-based and region-based algorithms which discriminate the focused condition
at the pixel, block or region level respectively.
However, the pixel-based methods such as \cite{image-partition-based}
suffer from misregistration \cite{Survey_SOTA}.
The block-based techniques such as \cite{old_SF} \cite{DSIFT} generally are sensitive
to the block-size \cite{Quadtree_structure}.
And the efficiency and performance of the region-based 
algorithms such as \cite{PCNN} \cite{BF} are usually influenced by the 
region segmentation procedures.


In the past few years, Deep Learning (DL) has aroused researchers
widespread interests for its surprising effectiveness in CV applications. 
Liu et al. \cite{CNN} made the first attempt to apply convolutional 
neural networks (CNNs) to MFIF.
In their work, the siamese architecture was used to extract the feature of the
focused and defocused regions.
Guo et al. \cite{FD-FCN} proposed a fully convolutional network for 
focus detection. And a cascaded boundary aware convolutional network \cite{MMF_NET}
was introduced to achieve clear results around the FDB.
Inspired by the DL tools and region based methods,
deep semantic segmentation and edge detection algorithms are employed 
in MFIF to capture clear focus maps
\cite{Boundary_aware, Image_Segmentation_Based}.

\begin{figure}[htbp]
    \centering
    \subfigure[$I_{A}$]{
    \includegraphics[scale = 0.15]{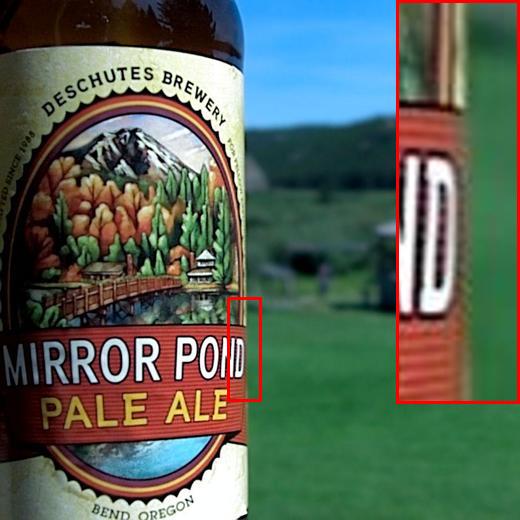}
    \label{Fig_detail_dse_Ia}}
    \subfigure[$I_{B}$]{
    \includegraphics[scale = 0.15]{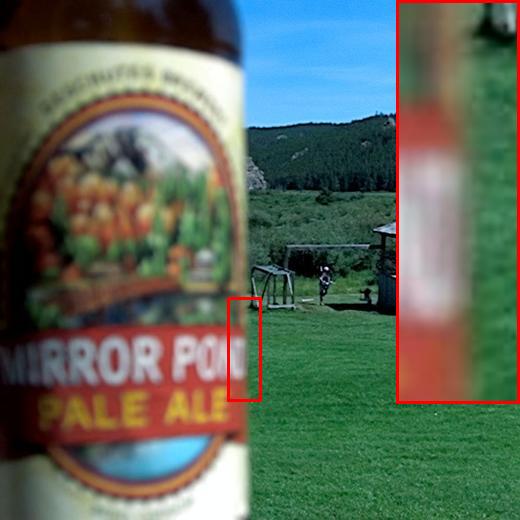}
    \label{Fig_detail_dse_Ib}}
    \caption{\footnotesize $I_{A}$ without DSE and $I_{B}$ suffering from DSE}
\end{figure}
  
However, apart from \cite{MMF_NET, MFFW}, few of previous works take the DSE into account.
So at present, in order to get fusion images with higher quality, 
one of the important research trends of MFIF is to aviod this spread effect.
Actually, the DSE describes a common phenomenon that the FDB sometimes is a ribbon 
region with the uncertain width instead of a clear curve.
More precisely, when the foreground is in focus,
the blurred background object will not influence the clear foreground region, 
that is, the FDB is clear as displayed in Fig.\ref{Fig_detail_dse_Ia} $I_{A}$. 
However, when the foreground is defocused, the blurred
foreground will permeate the background. Obviously, the FDB is not a 
clear line and blurred foreground objects are mildly bigger than 
itself as shown in Fig.\ref{Fig_detail_dse_Ib} $I_{B}$.

In this paper, we present a new DL-based MFIF algorithm termed MFIF-GAN 
to attenuate the DSE by generating focus maps in which the foreground regions 
are appropriately larger than the corresponding objects.
Specifically, the SE-ResNet \cite{SENet} is exploited as attention machanism 
and the reconstruction loss along with gradient penalty are utilized
to enhance the boundary details and improve the quality of fused results.
In addition, a large-scale training dataset suffering from the DSE is
synthesized by applying an $\alpha$-matte boundary defocus model 
\cite{MMF_NET} to the VOC 2012 dataset \cite{VOC2012}. 
At last, initial focus maps generated by the network are refined by
a post-processing algorithm in a computational efficient manner. 

A series of experiments are carried out to demonstrate the superiority of our method over
the state-of-the-art (SOTA) algorithms qualitively and quantitatively.
In order to verify the rationality of the generated focus maps,
we proposed a diffusion and contraction module to expand or shrink the foreground 
region at the pixel level.
In addition, ablation experiments are conducted to study 
the role of each element in our network. 

The contributions of this paper are as follows:
\begin{enumerate}
  \item A new multi-focus image fusion algorithm named MFIF-GAN
  with a new structure and well-designed loss functions is proposed.

  \item A more realistic dataset with DSE is constructed using
  an $\alpha$-matte model which can be a new benchmark training set
  for other supervised MFIF algorithms.

  \item Training on this new dataset makes the proposed network 
  a better performance qualitively and quantitatively than several SOTAs.
  With the computationally lightweight post-processing, the fusion procedure of 
  this algorithm is fastest with respect to the above methods.

  \item Generating focus maps with larger foregrounds could be a simple 
  and ingenious solution for further researches to alleviating the DSE. And a diffusion and contraction 
  module is firstly introduced to verify this statement.
\end{enumerate}

The rest of this paper is arranged as follows. In section
\ref{Related Works}, a briefly review of
related works is provided in which the $\alpha$-matte model,  
attention mechanisms and FuseGAN are introduced.
Section \ref{The Proposed Method} describes the details of the proposed network.
Then, contrast experiments are conducted in section \ref{Experiments} to evaluate 
all methods qualitatively and quantitatively.
The effectiveness of the new solution for attenuating the DSE is also
proved by diffusion and contraction experiments.
And ablation experiments are conducted in this section.
At last, conclusions are drawn in section \ref{Conclusions}.

\section{Related Works}
\label{Related Works}
\subsection{$\alpha$-matte Model for MFIF Datasets}

Due to the lack of large scale datasets of multi-focus images, several data 
generation methods based on public natural image datasets were adopted in many 
DL-based algorithms
\cite{fusegan, CNN, pixel_level, MSD_GLD, SESF}.
For example, in FuseGAN
\cite{fusegan}, 
a multi-focus image dataset was synthesized based
on PASCAL VOC 2012 dataset
\cite{VOC2012}. Fidel et al.\cite{FD-FCN} and Guo et al.\cite{Hourglass} 
used MS COCO and CIFAR-10 respectively to constructed MFIF training datasets.

However, the DSE is neglected in all datasets above.
The unrealistic training data may limit the performance of theses algorithms
\cite{MMF_NET}.
Therefore, Ma et al.
\cite{MMF_NET} proposed a novel $\alpha$-matte model which provides 
a insightful point of view to understand the DSE and real world multi-focus images. 
More details about the generation of training data is discussed in 
section \ref{dataset_generation}.


\subsection{Attention Mechanisms and Squeeze-Excitation Block}

Apart from traditional CNNs, some researchers attend to strengthen
the representation of networks to focus on salient objects in images 
for particular tasks. That is, in MFIF issue, consistent with the 
procedure of generating a rational focus map characterizing the 
objects which are in or out of focus.

As an attention mechanism, Hu et al. proposed the 
SE block \cite{SENet}
comsisting of a \textit{squeeze} and a \textit{excitation} operation, 
which model the interdependencies between 
the channels of feature maps to recalibrate them.
The squeeze module outputs a global distribution of features 
by aggregating feature maps across spatial dimensions.
And using a gating mechanism, the excitation operation produces a collection of weights 
representing the relationships between the channels.
Meanwhile, the SE block could be directly intergrated 
into other networks such as residual \cite{ResNet} 
and inception \cite{Inception_networks} networks as an atomic building block.
In our work, the combined SE-ResNet module is exploited
to extract the implicit features with multi-channels.

\subsection{FuseGAN for Multi-Focus Images to Focus Map Translation}
Inspired by the conditional generative adversarial network \cite{cGAN}
for image-to-image translation and the siamese network for extracting features 
of multi-focus images\cite{CNN}, FuseGAN \cite{fusegan} was proposed 
for MFIF task, which employed the objective function of LSGAN \cite{LSGAN} 
and exploited the convolutional conditional random fields (ConvCRFs) based 
technique \cite{ConvCRF} as a post-processing algorithm.
Compared with other spatial domain methods (CNN \cite{CNN}, BF \cite{BF} 
and DSIFT \cite{DSIFT}), the focused regions detected by FuseGAN are closer
to the ground truth.
In our experiments, FuseGAN is regarded as a baseline network.

The disadvantages of FuseGAN are summarized as follows. 
(1) As it is designed for gray images, the important color information may get lost,
that could limit the performance of the method. 
(2) It should be noted that the distinction 
between images will be excessively magnified by $\ell_2$-norm based
adversarial loss used in FuseGAN, which could make the 
training unstable. 
(3) As for the additional reconstruction loss function,
the coefficient $\lambda_{rec}$ of the binary cross entropy (BCE) loss
is set very large which has no interpretability.
And Aritra Ghosh et al. \cite{Robust_loss_function} argued that 
the cross entropy is commmonly sensitive to label noise in classification tasks.
(4) Last but not least, the ConvCRFs does not work
if all-in-focus images as ground truth are not available,
leading to unsatisfactory results in the real application. 


\section{The Proposed Method}
\label{The Proposed Method}
Lots of previous MFIF algorithms have achieved good fusion performances, but
few works concentrate on the DSE.
For example, the unsupervised MFF-GAN \cite{MFF-GAN}
which has a well-designed desision block and adaptive content loss function
does not analyse the essence of the defocus spreading.
In this section, we propose a supervised GAN-based network for the MFIF task,
especially for alleviating the DSE.
To begin with, we introduce the symbols used in this paper. 

The source image $I_A$ corresponds to the image which has a clear foreground 
with a blurred background. While another source $I_B$ suffering from the DSE 
has a blurred foreground with a clear background.
The all-in-focus image denotes as $I$. In the synthetic training 
set, it is regarded as the ground truth. The binary segmentation map $F$ 
represents a focus map, where $F_{ij}=1$ if $(i,j)$ pixel is in focus and 0 otherwise. 
The focus map generated by the network and the one refined by post-processing
are denoted by $\hat{F}$ and $\hat{F}_{final}$ respectively.


\subsection{Architecture of the Network}

\begin{figure}[H]
  \centering
  \includegraphics[scale = 0.66]{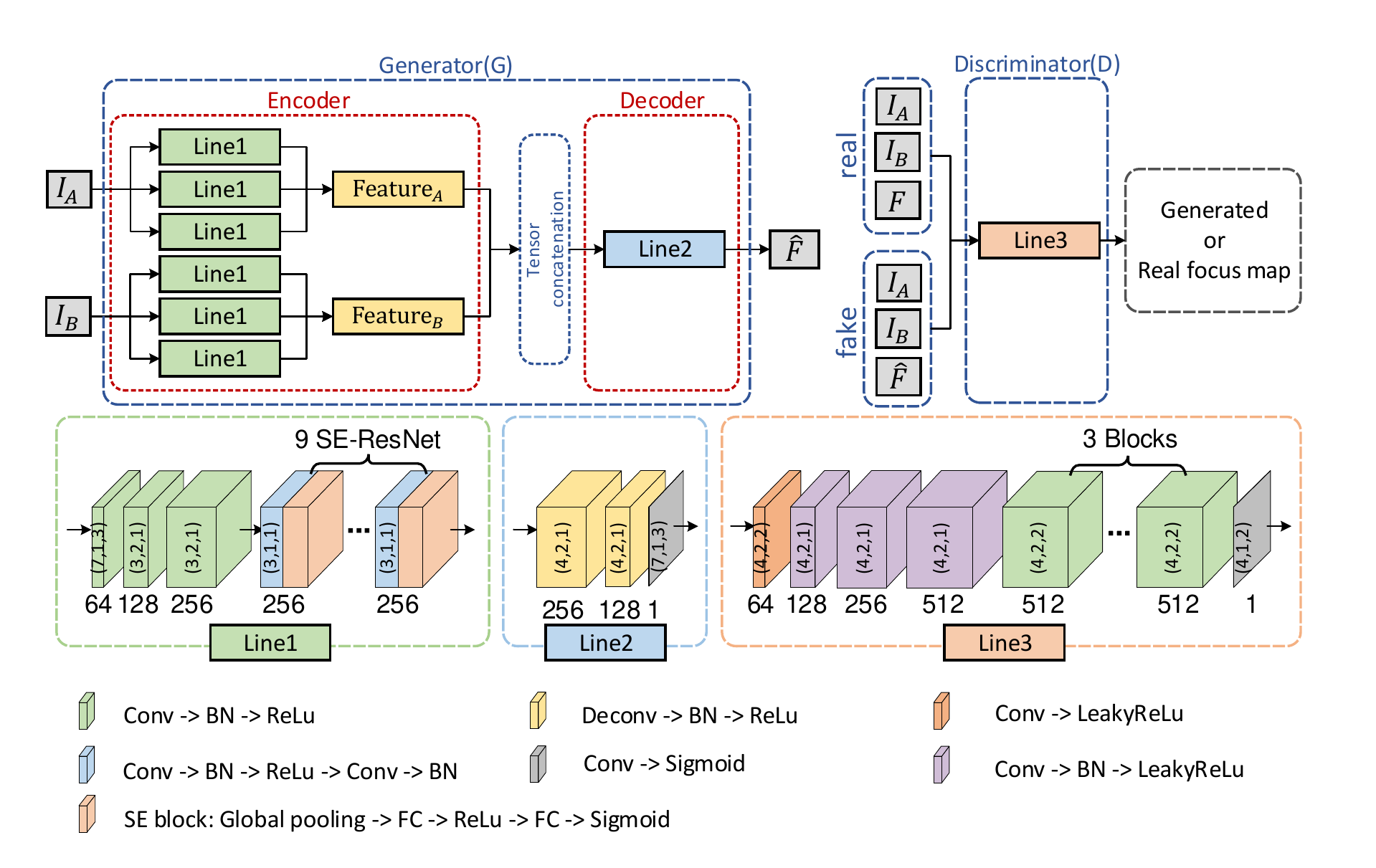}
  \caption{\footnotesize The work flow of the training and the detail architecture of MFIF-GAN.
  The convolutional layer, transposed convolutional layer, 
  BatchNorm layer, Rectified Linear Unit and fully connected layer are denoted as 
  Conv, Deconv, BN, ReLU and FC respectively.
  The number under every block represents the channel number of Conv or Deconv.
  The array in bracket in every block indicates the kernel, stride and
  padding size of Conv or Deconv respectively.}
  \label{Architecture}
\end{figure}

As a variant of GAN \cite{GAN}, MFIF-GAN also consists of two fundamental modules: a generator and a 
discriminator.
The architecture of our network is shown in Fig.\ref{Architecture}.
The generator in MFIF-GAN is fed with the source color images $I_A$ and $I_B$
aiming to generate focus map $\hat{F}$. The inputs of the discriminator 
are the concatenation of $I_A$, $I_B$ and the (real or generated) focus map. 
The aim of the generator is to reconstruct focus maps as accurately as possible, 
while the purpose of the discriminator is to distinguish the generated 
focus maps from the real ones.

\textbf{Generator $G$:}
The $G$ includes an encoder, a tensor concatenation module and a decoder. 
In order to effectively process color images, the encoder is designed as six branches of 
parallel sub-networks sharing parameters for each channel of source images.

Each sub-network in encoder is composed of 
three convolutional modules and nine residual blocks. 
In order to reduce computation burden, the 2nd and 3rd convolutional modules 
with a stride of 2 down-sample feature maps. 
Furthermore, inspired by SE-Net \cite{SENet} as an attention mechanism, 
each residual block is equipped with a SE block as SE-ResNet
to find the interdependencies between the channels of feature maps 
and extract the most informative components of the images.
In every SE-ResNet module, the SE block is attached to the non-identity branch
which is a defaulted residual module with two convolutional layers and 
batch normalization (BatchNorm) \cite{Batchnorm} to guarantee the squeeze and 
excitation operations work before addition with the identity branch.

In the tensor concatenation part, six feature maps 
extracted by the encoder from each channel are
averaged to obtain the global features $Feature_{A}$, $Feature_{B}$ 
of $I_{A}$ and $I_{B}$, respectively.
Then, $Feature_{A}$ and $Feature_{B}$ are concatenated on the third channel.

In the decoder, the joint feature is upsampled and deconvolved 
through two transposed convolutional layers for size recovery and reconstruction. 
Finally, the single-channel focus map is output by a convolutional
and activating layer.

\textbf{Discriminator $D$:}
In the discriminator, eight convolutional 
layers are used to compress the input data continuously until the final 
sigmoid activation function is used to judge whether the input focus map is real 
or generated.
Specifically, the input is a 7-channel tensor, i.e., the concatenation of source images 
and a focus map which is generated by $G$ or be the real one from the 
training dataset. The kernal, padding and stride size of convolutional layers for 
down-sampling are set to guarantee the final output is a single element for judgment.


\subsection{Loss Function}
The objective function plays a crucial role in deep learning. Some works indicated
that the original GAN suffers from training instability because 
during the $D$ and $G$ are trained to
optimality alternately, the Jensen-Shannon divergence between the real 
data distribution $\mathbb{P}_{data}$ 
and the noise distribution $\mathbb{P}_{z}$ is minimized \cite{GAN}, which often 
leads to vanishing gradients.
As a improved version, WGAN \cite{WGAN} is still subjected to either vanishing or exploding
gradients without rational tuning of the clipping threshold $c$ \cite{improved_WGAN}.

So in our work, we adopted the improved training of WGAN \cite{improved_WGAN},
the adversarial loss function of $D$ and $G$ are listed as equations
(\ref{adv_D}) and (\ref{adv_G}):
\begin{equation}
    \mathcal{L}_{adv}(D)=\mathbb{E}_{I_{A,B}\thicksim 
    \mathbb{P}_{I_{A,B}}}[D(I_{A,B},G(I_{A,B}))]-
    \mathbb{E}_{I_{A,B},F\thicksim \mathbb{P}_{I_{A,B},F}}[D(I_{A,B},F)]
    \label{adv_D}
\end{equation}
\begin{equation}
    \mathcal{L}_{adv}(G)=-\mathbb{E}_{I_{A,B}\thicksim 
    \mathbb{P}_{I_{A,B}}}[D(I_{A,B},G(I_{A,B}))]
    \label{adv_G}
\end{equation}
where $I_{A,B},F\thicksim \mathbb{P}_{I_{A,B},F}$ in (\ref{adv_D}) 
denotes the inputs of $D$ follow the joint distribution of the 
couple of images and focus maps from the real data.

As the main contribution in \cite{improved_WGAN},
the gradient penalty $\mathcal{L}_{gp}$ is added into the loss of 
$D$ to stabilize the training process
and further improve the quality of generated focus maps. That is:
\begin{equation}
    \mathcal{L}_{gp}=\mathbb{E}_{I_{A,B},\tilde{F}\thicksim 
    \mathbb{P}_{I_{A,B},\tilde{F}}}
    [(\|\bigtriangledown_{\tilde{F}}D(I_{A,B},\tilde{F})\|_{2}-1)^{2}]
    \label{gradient_penalty}
\end{equation}
where $\tilde{F}$ is sampled uniformly along a straight line between $F$ and $\hat{F}$.

According to \cite{Robust_loss_function}, 
compared with cross entropy loss used in FuseGAN, the loss function based on
mean absolute value of error is more robust to the noise.
Therefore, the $\ell_1$-norm is utilized as reconstruction 
loss $\mathcal{L}_{rec}$ to measure the 
difference between the generated focus maps and the real ones, 
as shown in equation (\ref{reconstruction}):
\begin{equation}
    \mathcal{L}_{rec}=\mathbb{E}_{I_{A,B},F\thicksim 
    \mathbb{P}_{I_{A,B},F}}[|F-G(I_{A,B})|]
    \label{reconstruction}
\end{equation}

So the total loss functions of MFIF-GAN can be defined
by (\ref{total_D}) and (\ref{total_G}):
\begin{equation}
    \mathop{\min}_{D}\mathcal{L}(D)=\mathcal{L}_{adv}(D)+\lambda_{gp}\mathcal{L}_{gp}
    \label{total_D}
\end{equation}
\begin{equation}
    \mathop{\min}_{G}\mathcal{L}(G)=\mathcal{L}_{adv}(G)+\lambda_{rec}\mathcal{L}_{rec}
    \label{total_G}
\end{equation}
We use $\lambda_{gp}=10$ defaulted in \cite{improved_WGAN} for all experiments.
In order that $\lambda_{gp}$ and $\lambda_{rec}$ are used to adjust these two
additional loss terms to the same level of importance,
the value of $\lambda_{rec}$ is set as same as $\lambda_{gp}$.


\subsection{Post-processing in MFIF-GAN}

The focus maps generated by $G$ often suffer from
misregistration or noise resulting in unsatisfactory fusion images.
Therefore, we employ the small region removal (SRR) strategy for refinement. The SRR works on the binary matrix and it
removes the region whose number of pixels is smaller than a threshold $N$.
In this paper, we set $N=0.001WH$, where $W$ and $H$ are the 
width and height of an image, respectively. The final 
focus maps $\hat{F}_{final}$ can be obtained after this post-processing
which is so simple and effective that it does not increase computational burden.


\section{Experiments}
\label{Experiments}
In this section, the preparations of experiments including datasets,
assessment metrics along with setting of training and testing
are firstly introduced. 
Based on three public test datasets, we compare our MFIF-GAN with
eight representative SOTA methods, including  
spetial domain methods Quadtree \cite{Quadtree}
\footnote{\url{https://github.com/uzeful/Quadtree-Based-Multi-focus-Image-Fusion}}
and DSIFT \cite{DSIFT}
\footnote{\url{http://www.escience.cn/people/liuyu1/Codes.html}}, 
transform domain methods NSCT \cite{NSCT}
\footnote{\url{https://github.com/yuliu316316/MST-SR-Fusion-Toolbox}}, CSR \cite{CSR}
\footnote{\url{http://www.escience.cn/people/liuyu1/Codes.html}} and MWGF \cite{MWGF}
\footnote{\url{https://www.researchgate.net/publication/307415978
\_MATLAB\_Code\_of\_Our\_Multi-focus\_Image\_Fusion\_Algorithm\_MWGF}},
deep learning based methods MMF-Net \cite{MMF_NET}
\footnote{\url{https://github.com/xytmhy/MMF-Net-Multi-Focus-Image-Fusion}},
CNN \cite{CNN}
\footnote{\url{http://home.ustc.edu.cn/~liuyu1}}
and FuseGAN \cite{fusegan}
\footnote{The official codes of FuseGAN are unavailable, so we 
re-implement and re-train FuseGAN.}.
Qualitative and quantitative results of all methods are provided
in detail. In addition, the difficulty of avioding the DSE is analysed and a simple and 
effective solution is put forward. Subsequently, the edge diffusion and 
contraction experiments are proposed to verify the rationality of this solution.
Finally, the ablation experiments are conducted to validate the 
contributions of several modules in our method.


\subsection{Experiments Setting}

\subsubsection{Dataset}
\label{dataset_generation}
The MFIF training datasets which take the DSE into account are not 
available publicly. 
Therefore, we apply the $\alpha$-matte model \cite{MMF_NET}
to the PASCAL VOC 2012 \cite{VOC2012} 
to construct a synthetic training dataset with DSE which will 
be termed $\alpha$-matte dataset.

Each picture in this  image segmentation database 
PASCAL VOC 2012 is accompanied by a segmentation map. 
We regard the binary segmentation map as a focus map $F$ 
(matte $\alpha^{C}$ in \cite{MMF_NET}).
Using $F$, 
the clear foreground $FG^{C}$ and background $BG^{C}$ can be got
as follows:
\begin{equation}
    FG^{C}=F*I
    \label{clear_foreground}
\end{equation}
\begin{equation}
    BG^{C}=(1-F)*I
    \label{clear_background}
\end{equation}
where $*$ means pixel-wise production.

The blurred focus map $F^{B}$ (matte $\alpha^{B}$ in \cite{MMF_NET}) can 
be obtained by applying a gaussian filter $G(x,y;\sigma)$ 
to corresponding $F$. That is:
\begin{equation}
    F^{B}=G(x,y;\sigma)\otimes F
    \label{blurred_focus_map}
\end{equation}
where $\otimes$ represents the convolutional operator. 
The blurred foreground $FG^{B}$ and background $BG^{B}$ can be acquired in the 
same way.

Finally, according to the $\alpha$-matte model, a pair of training 
images $I_{A}$ and $I_{B}$ with only two 
valid surface (foreground surface $S_{FG}$ and background surface $S_{BG}$)
can be obtained by equations (\ref{I_A}) and (\ref{I_B}) respectively.
\begin{equation}
  \begin{split}
    I_{A}&=S_{FG}^{clear}+S_{BG}^{blurry}\\
         &=FG^{C}+(1-\alpha^{C})*BG^{B}\\
         &=F*I+(1-F)* \left\{ G(x,y;\sigma)\otimes \left[ (1-F)*I \right] \right\}\\
  \end{split}
  \label{I_A}
\end{equation}
\begin{equation}
  \begin{split}
         I_{B}&=S_{FG}^{blurry}+S_{BG}^{clear}\\
         &=FG^{B}+(1-\alpha^{B})*BG^{C}\\
         &=G(x,y;\sigma)\otimes \left( F*I \right) +(1-G(x,y;\sigma)\otimes F)* \left[ (1-F)*I \right]
  \end{split}
  \label{I_B}
\end{equation}

In order to verify the contribution of the $\alpha$-matte model, 
based on the generation method raised in \cite{Hourglass}, an another dataset 
without DSE is synthesized for the ablation experiment. 
In what follows, we call it conventional 
MFIF training dataset. In formula, the source images are obtained by
(\ref{conventional_I_A}) (\ref{conventional_I_B}).
\begin{equation}
    I_{A}=F*I+(1-F)*(G(x,y;\sigma)\otimes I)
    \label{conventional_I_A}
\end{equation}
\begin{equation}
    I_{B}=F*(G(x,y;\sigma)\otimes I)+(1-F)*I
    \label{conventional_I_B}
\end{equation}

As for testing data, the famous Lytro \cite{Lytro} dataset
is utilized.
In addition, a new dataset called MFFW \cite{MFFW}
which significantly suffers from the DSE is employed in the test. 
In order to verify the performance of our algorithm comprehensively, 10 pairs of gray images 
termed grayscale \cite{Survey_SOTA} is also used.

 \subsubsection{Training and Testing Setup}

 In the training stage, we optimize $G$ and $D$ alternately.
 In order to better optimize the objective function and simplify the updating strategy 
 of learning rate, we use the adam with two parameters 
 $\beta_{1}$ and $\beta_{2}$ which are initialized to 
 0.5 and 0.999, respectively. And the linear declining strategy is used to update 
 the learning rates of $G$ and $D$ both initialized to $0.0001$.
 Besides, the update rate ratio between $G$ and $D$ is $1:5$, which means that $G$ is updated once 
 after updating $D$ for five times. 

 In the testing phase, we only retain $G$ followed by a SRR to generate the 
 focus maps $\hat{F}$ and refine them. The processed focus maps $\hat{F}_{final}$
 are used to extract the clear regions and reconstruct the all-in-focus images 
 as follows:
\begin{equation}
    I_{fused}=I_{A}*\hat{F}_{final} + I_{B}*(1-\hat{F}_{final})
    \label{fuse_equation}
\end{equation}

 For the grayscale dataset, samples are tripled
 to form images with 3 channels as inputs of $G$.

 \subsubsection{Quantitative Assessment Metrics}
 
 In order to evaluate the performance of different algorithms 
 comprehensively, twelve objective metrics are utilized
 \footnote{The implementation of these metrics are available at 
 \url{https://github.com/zhengliu6699/imageFusionMetrics}},
 which are (1) information theory-based metrics including mutual information $MI$
 \cite{MI}, Tsallis entropy based metric $TE$
 \cite{TE} and nonlinear correlation information entropy $NCIE$ \cite{NCIE};
 (2) image structure similarity-based metric: 
 structure similarity index measure ($SSIM$) based metric proposedal by Cui Yang et al.
 (also named Yang's metric $Q_{Y}$) \cite{Q_Y};
 (3) human perception inspired fusion metric: Chen-blum metric $Q_{CB}$
 \cite{CB}
 (4) image feature-based metrics including gradient-based metric $Q_{G}$ \cite{Q_G},
 multiscale scheme based $Q_{M}$ \cite{Q_M},
 spatial frequency based $SF$ \cite{SF},
 linear index of fuzziness $LIF$ \cite{LIF},
 average gradient $AG$\cite{AG},
 mean square deviation $MSD$ \cite{MSD_GLD} 
 and gray level difference $GLD$ \cite{MSD_GLD}.
 Detailed mathematical expressions of these metrics could be found in original papers
 or overview work \cite{metrics_analyses}.
 
 It is worth to note that the fused images are better if all metrics are larger 
 except for $LIF$.
 These metrics have different emphases, so none of them is better than all others.
 The first eight traditional metrics (i.e. $MI$, $TE$, $NCIE$, $Q_{Y}$,
 $Q_{CB}$, $Q_{G}$, $Q_{M}$ and $SF$) are widely used in assessment of images
 quality for their characteristics of computing agreements of fused
 images with the sources. Instead, the last four metrics 
 (i.e. $LIF$, $AG$, $MSD$ and $GLD$) are used to evaluate
 the performance of edge detail and contrast enhancement of the fused
 results in spite of the sources \cite{MSD_GLD}.
 
 \subsection{Comparison with SOTA Methods}

 \subsubsection{Quantitative Comparison}
 Based on the above test datasets and evaluation metrics, 
 the quantitative results with respect to the proposed MFIF-GAN and 
 SOTAs are listed in Tab.\ref{table_compare_sota}.

 On the Lytro dataset, it can be clearly seen that 
 on the first eight traditional metrics, our proposed MFIF-GAN 
 trained on the $\alpha$-matte dataset is generally superior to other methods. 
 Moreover, on the last four metrics which evaluate the edge quality, 
 MFIF-GAN can still take the lead 
 in addition to the MMF-Net which is specially designed 
 and optimized for the DSE.
 On the MFFW, the absolute advantage of our MFIF-GAN compared with other SOTAs 
 is obvious. As for the grayscale dataset, our method achieves comparable results 
 to the Quadtree generally.

\begin{table*}[htbp]
    \small
    \scriptsize
    \centering
    \caption{\footnotesize Average scores of fusion result based on Lytro, MFFW and grayscale datasets by 
    all algorithms on 12 metrics. The best, the second best, 
    and the third best results are highlighted in bold, double underlining, and underlining, respectively.}
    \resizebox{\textwidth}{!}{
    \subtable[Lytro]{
    \begin{tabular}{cccccccccc}
    \toprule
       ~& CNN & MMF-net & MWGF & Quadtree  & DSIFT & CSR & NSCT & FuseGAN & MFIF-GAN\\
    \midrule
      $MI$    &\underline{1.07075}&0.92506 &1.01685 &1.05303  &\uuline{1.08438} & 0.99020 & 0.90903&1.05501 &\textbf{1.09446}\\
  
      $TE$    &\underline{0.37853}&0.36443 &0.37221 & 0.37658 &\uuline{0.37925} & 0.37288 & 0.36594    & 0.37610 &\textbf{0.38034} \\
  
      $NCIE$  &\underline{0.83933}&0.83067 &0.83603 & 0.83806& \uuline{0.84021} & 0.83400 & 0.82957    &0.83896   &\textbf{0.84097}\\
  
      $Q_{G}$ &0.70763& 0.64492&\uuline{0.71059} &0.69854 & 0.70118& 0.69508 & 0.68305    & \underline{0.70814} &\textbf{0.71786}\\
  
      $Q_{M}$ &\underline{1.91707}& 1.42079& 1.73044&1.87318  &\uuline{2.03527} & 1.63746 & 1.40236    &1.77586 &\textbf{2.07952}\\
  
      $SF$    &-0.03422& \textbf{-0.00845}&-0.03875 & -0.02546&\underline{-0.02442} &-0.03371 & -0.03258    & -0.03629 &\uuline{-0.02324}\\
  
      $Q_{Y}$ &\underline{0.97583}&0.94947 & 0.97004&0.973990 &\uuline{0.97615} & 0.95141 &0.9533    & 0.97419 &\textbf{0.97696}\\
  
      $Q_{CB}$&\underline{0.79612}&0.74312 & 0.77483& 0.78761& \textbf{0.79886}& 0.76064& 0.74455    & 0.784 &\uuline{0.79764}\\
  
      $LIF$   &0.38740&\textbf{0.38670} & 0.387790&0.38714 & \uuline{0.38694}& 0.38737 &0.38959    &0.38823  &\underline{0.38698}\\
  
      $AG$    &2.99603&\textbf{3.09201} &2.97081 & 3.02048& \underline{3.02212}& 2.97995 & 3.00927    &3.00098 &\uuline{3.0266}\\
  
      $MSD$   &0.07007& \textbf{0.07059}& 0.07000&0.070130 &\underline{ 0.07017}& 0.07001 &0.07003    & 0.06999  &\uuline{0.07019}\\
  
      $GLD$   &14.75660& \textbf{15.23405}&14.62293 &14.87703  & \underline{14.88406}& 14.67629 & 14.82605    & 14.77889  &\uuline{14.90642}\\
      \bottomrule
    \end{tabular}
    }}
    \resizebox{\textwidth}{!}{
    \subtable[MFFW]{
    \begin{tabular}{ccccccccc}
    \toprule
       ~& CNN & MWGF & Quadtree  & DSIFT & CSR & NSCT & FuseGAN  &  MFIF-GAN\\
    \midrule
      $MI$    &0.99738 &	0.96529	&\uuline{1.02036}&	0.98671&	0.8907  &	0.78236 &\underline{1.01877}&	\textbf{1.06806}   \\
  
      $TE$    &0.364&	0.35761	&\underline{0.36466}	&	0.35911&	0.3475& 0.34291  &	\uuline{	0.36908} &	\textbf{0.37169}    \\
  
      $NCIE$  &0.83288&	0.83086&	\underline{0.83451}&0.83298	&0.8271	&0.82123  &\uuline{0.83531}&	\textbf{0.83716 }   \\
  
      $Q_{G}$ &0.56866&	\underline{0.60566}	&0.49265&	\textbf{0.63777}	&0.53382 & 0.56998  &	\uuline{0.62866} &	0.56345    \\
  
      $Q_{M}$ &1.94031&	1.8218&	\uuline{2.08961}	&\underline{2.03046}	&1.88064	 &1.14658  &1.83548 &	\textbf{2.21075}    \\
      
      $SF$    &-0.05251&	-0.05472&	-0.04142&	\uuline{-0.03498}&	\underline{-0.03717} &-0.04652  &	-0.04685 &	\textbf{-0.03111}    \\
  
      $Q_{Y}$ &0.96754&	\underline{0.97207}&	0.96834	&0.93459&	0.86835&0.91066  &\uuline{0.97554}  &	\textbf{0.97939}    \\
  
      $Q_{CB}$&0.74025&	0.74085& \uuline{0.75113}&	0.72927	&0.69403 &0.67343&\underline{0.74159}&\textbf{0.75624}   \\
  
      $LIF$   &0.38654&	0.38852	&\underline{0.38617} & 0.38884&	0.38709 &	0.38626  &\uuline{0.38522} &	\textbf{0.38361}    \\
  
      $AG$    &3.53508&	3.51182	&3.60422&		\uuline{3.62413}	&\textbf{3.63075} &	3.59994 &	3.55498 &\underline{3.62057}    \\
  
      $MSD$   &\textbf{0.07931} &	0.07833	&0.07857&	0.07846&	0.07843& \uuline{0.07922}  &		0.07879  &	\underline{0.07902}    \\
  
      $GLD$   &17.49452	&17.36662&	17.83513&	\uuline{17.92287}&	\textbf{17.95869} &17.80467   &		17.59685  & \underline{17.91793 }\\
      \bottomrule
    \end{tabular}
    }}

  \resizebox{\textwidth}{!}{
  \subtable[grayscale]{
    \begin{tabular}{ccccccccc}
    \toprule
       ~& CNN & MWGF & Quadtree  & DSIFT & CSR & NSCT & FuseGAN  &  MFIF-GAN\\
    \midrule
      $MI$    &1.1222 &	1.10868	 & \textbf{1.15406} &\uuline{1.14503} & 	1.04652  & 0.85708  & 	1.07939 &	\underline{1.13731}    \\
  
      $TE$    &0.41359 &		0.41237	 & \uuline{0.41746 } & 	\underline{0.41653}	 & 0.41127   & 0.38899 &		0.41031	 &\textbf{0.41859}   \\
  
      $NCIE$  &0.8376 &		0.83731 & 	\textbf{0.83957}	& \underline{0.83889}	 & 0.83307  & 0.82378 &	0.83649 &	\uuline{0.83899 }   \\
  
      $Q_{G}$ &0.67711	 &\underline{	0.67974} & 	0.63484 &\uuline{ 0.68076} & 	0.67628   & 0.63163  &	0.61953	 &	\textbf{0.68343} \\
  
      $Q_{M}$ &2.33876  &		2.32259 & \textbf{2.46421} & \uuline{	2.44523 }& 	2.22342  & 1.55645  &	1.73248  &	\underline{2.3914} \\
      
      $SF$    &-0.0414 &-0.04321	 & \underline{-0.03422}& 	-0.0353	 & -0.03964  &-0.03928 &	\textbf{0.02467}&		\uuline{-0.03342}    \\
  
      $Q_{Y}$ &\underline{0.97724} & \uuline{0.97809	} & \textbf{0.97873}& 0.97517	 & 0.93837  &  0.93601  &	0.95378 &	0.97348    \\
  
      $Q_{CB}$&\underline{0.76327} &\textbf{0.76461} & 	0.76313 & \uuline{0.76427 }& 	0.72702   & 0.7029    &	0.74415  &		0.7558    \\
  
      $LIF$   &0.47571 &	0.47552 & \underline{0.47551}	& 	0.47556	 & 0.47578 & \uuline{0.47465} &	0.49394 &	\textbf{0.47458}   \\
  
      $AG$    &3.53597 &	3.53821 & \uuline{3.56883} & 	\underline{3.5687} & 	3.52606   &  \textbf{3.58797 }  &	3.54217 	 &\underline{	3.56084 }   \\
  
      $MSD$   &0.13997 &	0.13984 & 	0.14002	& 0.14009 & 	0.13987   & \underline{0.14014}  &	\textbf{0.15071}&	\uuline{0.14064 }   \\
  
      $GLD$   &17.15701	 &17.1702 & \uuline{17.31683} &\underline{17.31368 }&17.10948 &	\textbf{17.42544 }  & 17.1133&	\underline{17.27329  } \\
      \bottomrule
    \end{tabular}
  }}
    \label{table_compare_sota}
  \end{table*}

\subsubsection{Visual Comparison of Details}
\begin{figure*}[htbp]
  \centering
  \subfigure[CNN]{
  \includegraphics[scale = 0.12]{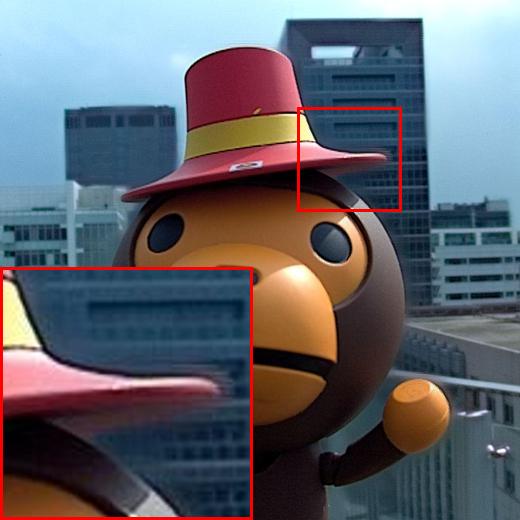}}
  \subfigure[MMF-net]{
  \includegraphics[scale = 0.12]{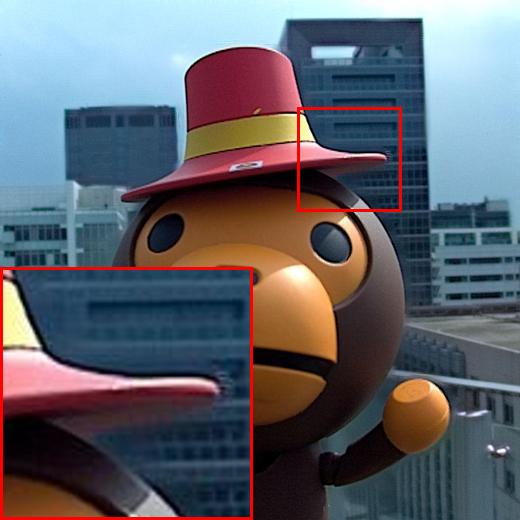}}
  \subfigure[MWGF]{
  \includegraphics[scale = 0.12]{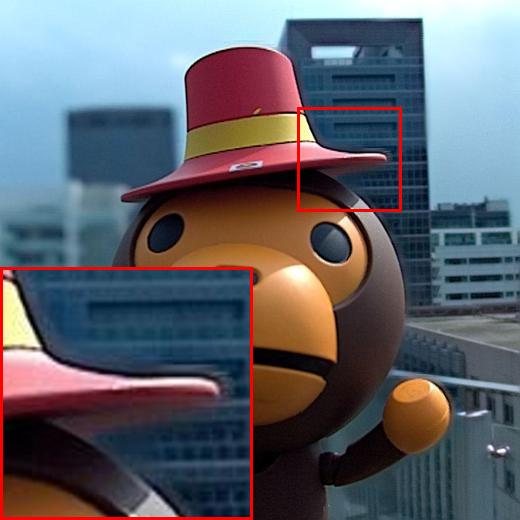}}
  \subfigure[Quadtree]{
  \includegraphics[scale = 0.12]{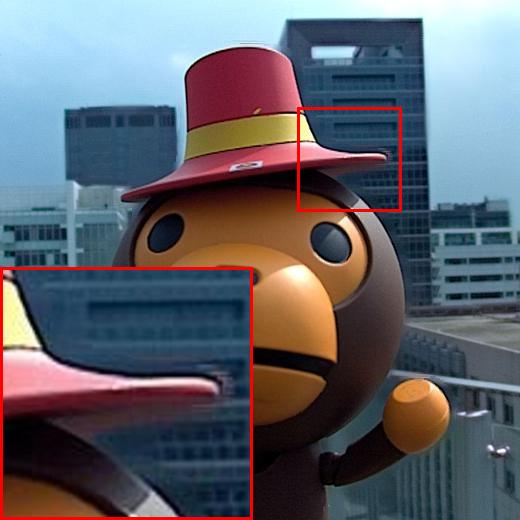}}
  \subfigure[BF]{
  \includegraphics[scale = 0.12]{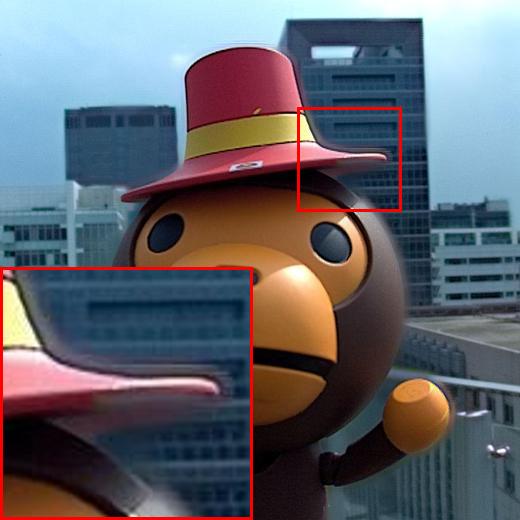}}
  \subfigure[DSIFT]{
  \includegraphics[scale = 0.12]{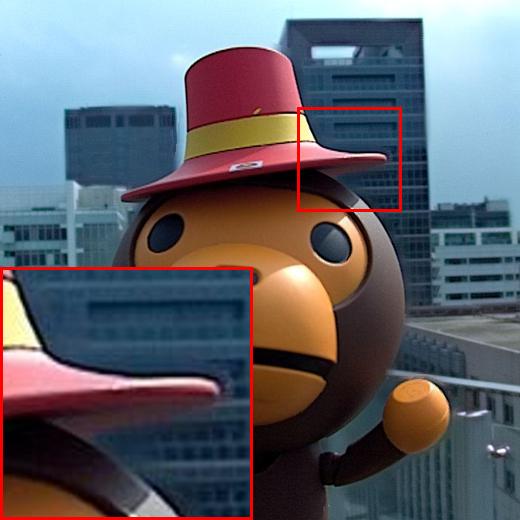}}
  \subfigure[CSR]{
  \includegraphics[scale = 0.12]{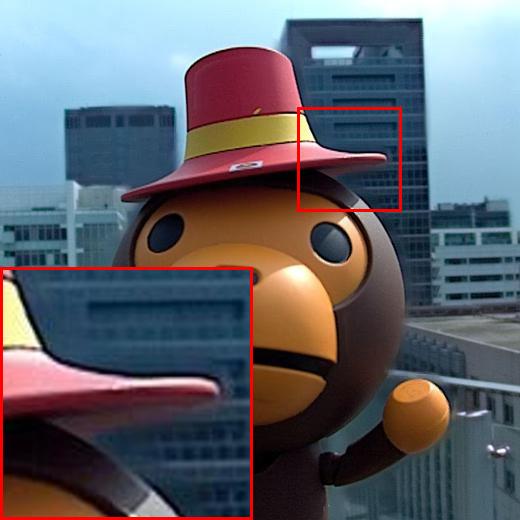}}
  \subfigure[FuseGAN]{
  \includegraphics[scale = 0.12]{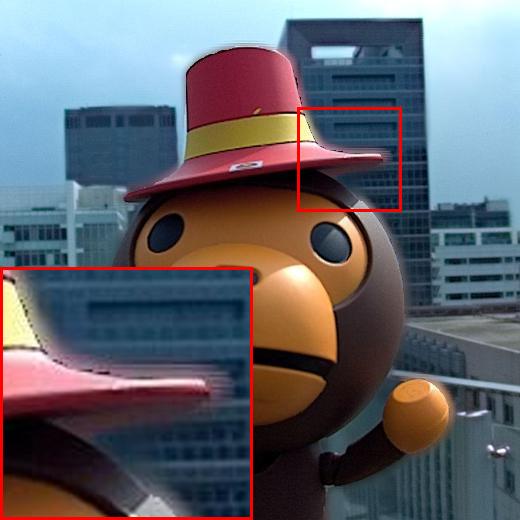}\label{Lytro20_FuseGAN}}
  \subfigure[FuseGAN\_Ib]{
  \includegraphics[scale = 0.12]{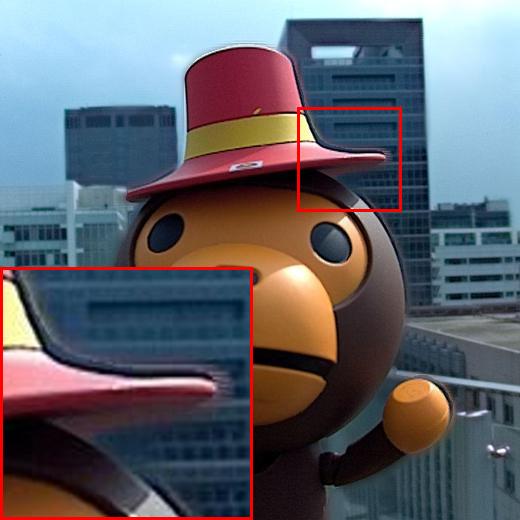}\label{Lytro20_FuseGAN_Ib}}
  \subfigure[MFIF-GAN]{
  \includegraphics[scale = 0.12]{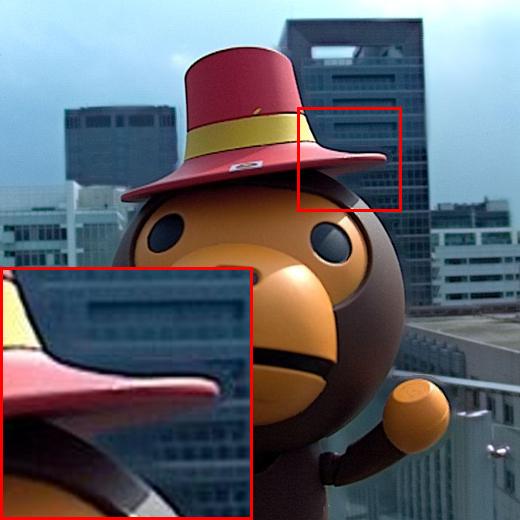}}
  \subfigure[CNN]{
  \includegraphics[scale = 0.11]{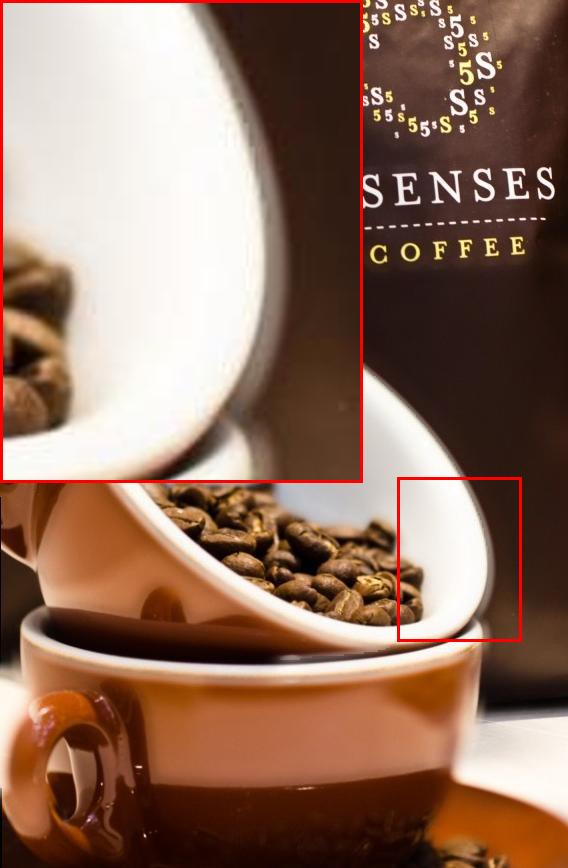}}
  \subfigure[MWGF]{
  \includegraphics[scale = 0.11]{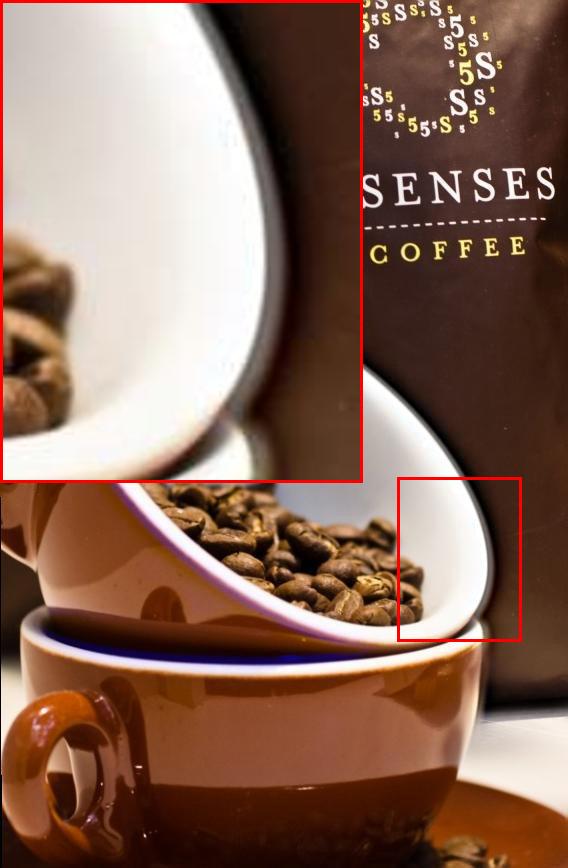}}
  \subfigure[Quadtree]{
  \includegraphics[scale = 0.11]{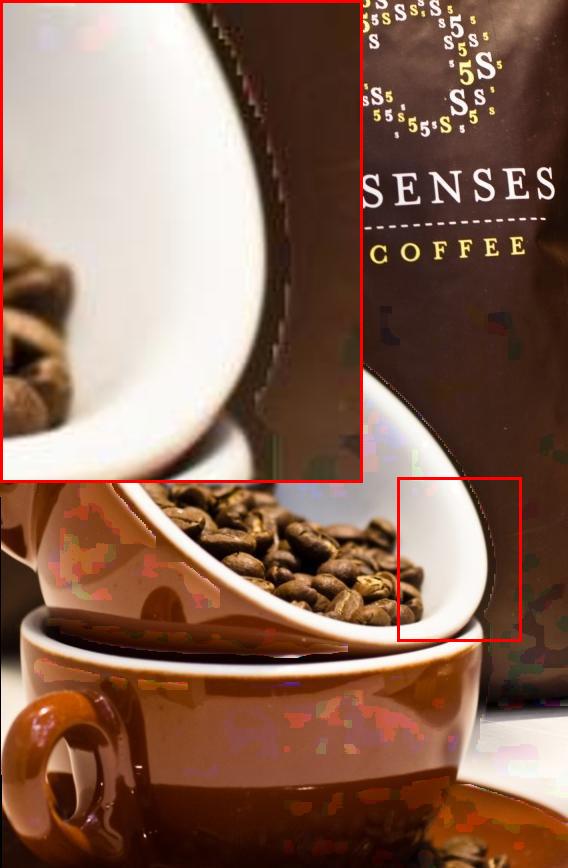}}
  \subfigure[BF]{
  \includegraphics[scale = 0.11]{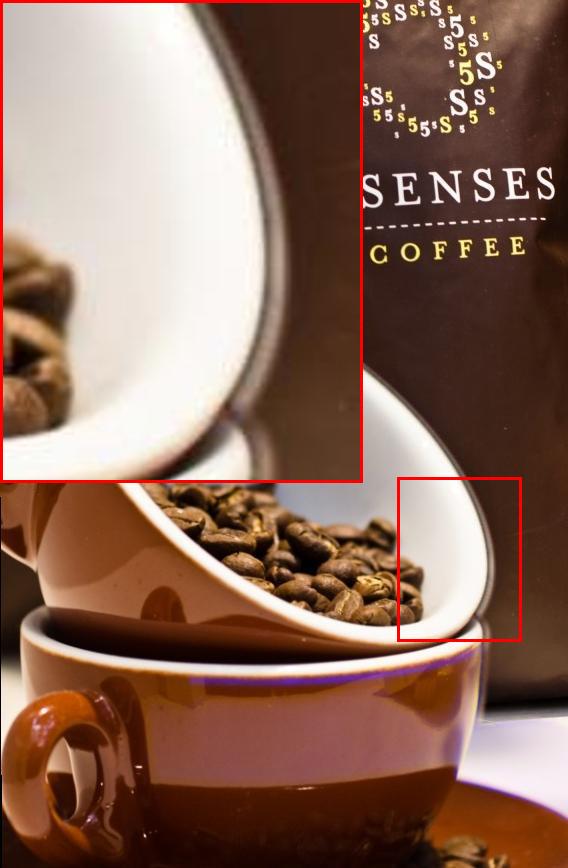}}
  \subfigure[DSIFT]{
  \includegraphics[scale = 0.11]{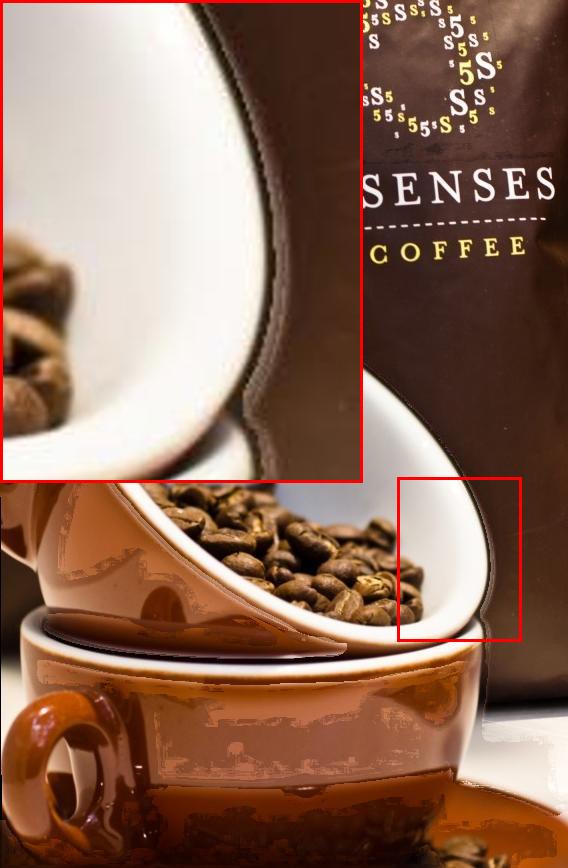}}
  \subfigure[CSR]{
  \includegraphics[scale = 0.11]{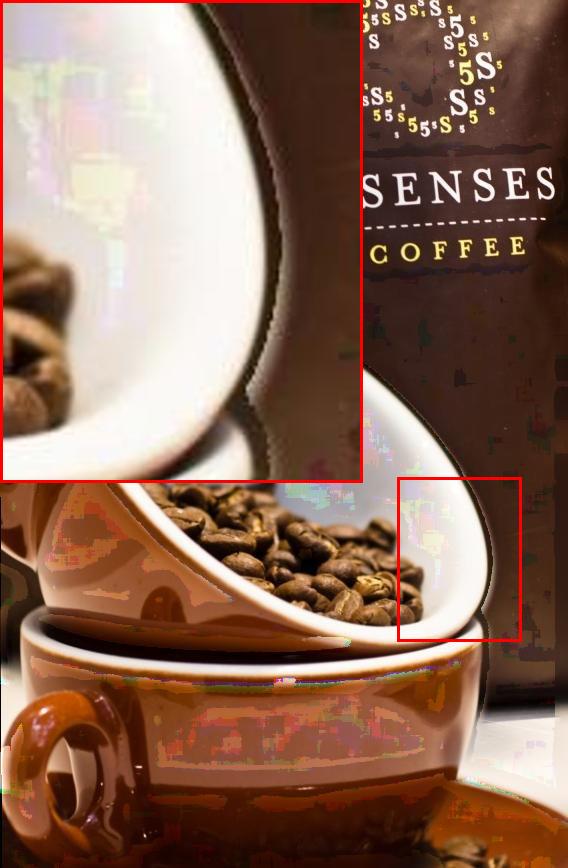}}
  \subfigure[FuseGAN]{
  \includegraphics[scale = 0.11]{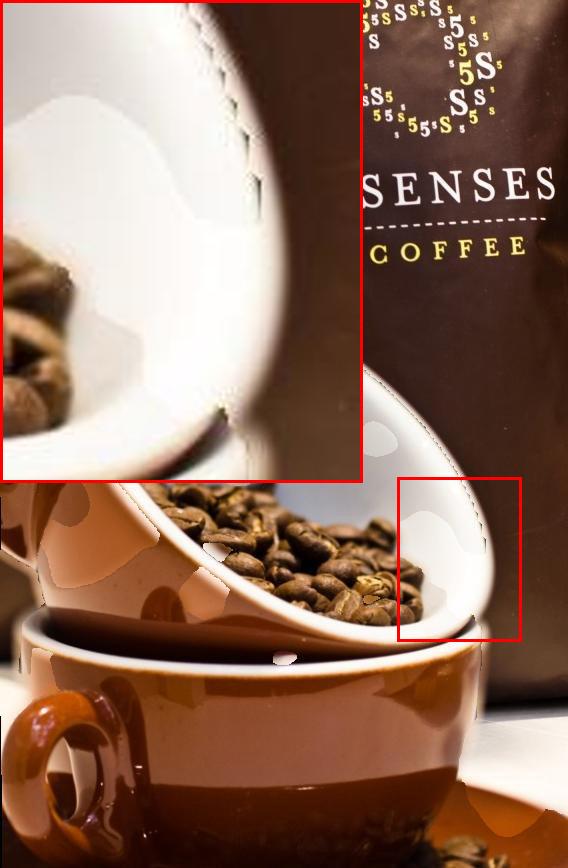}\label{MFFW11_FuseGAN}}
  \subfigure[FuseGAN\_Ib]{
  \includegraphics[scale = 0.11]{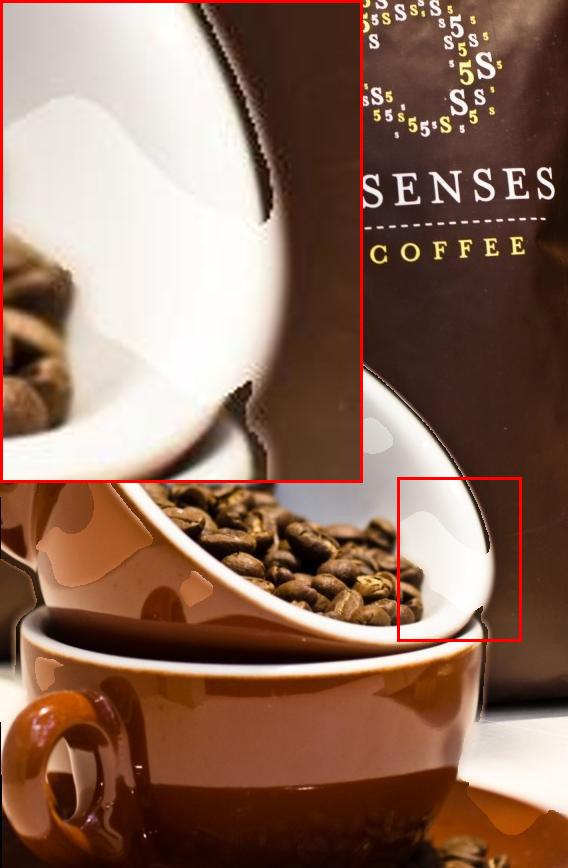}\label{MFFW11_FuseGAN_Ib}}
  \subfigure[MFIF-GAN]{
  \includegraphics[scale = 0.11]{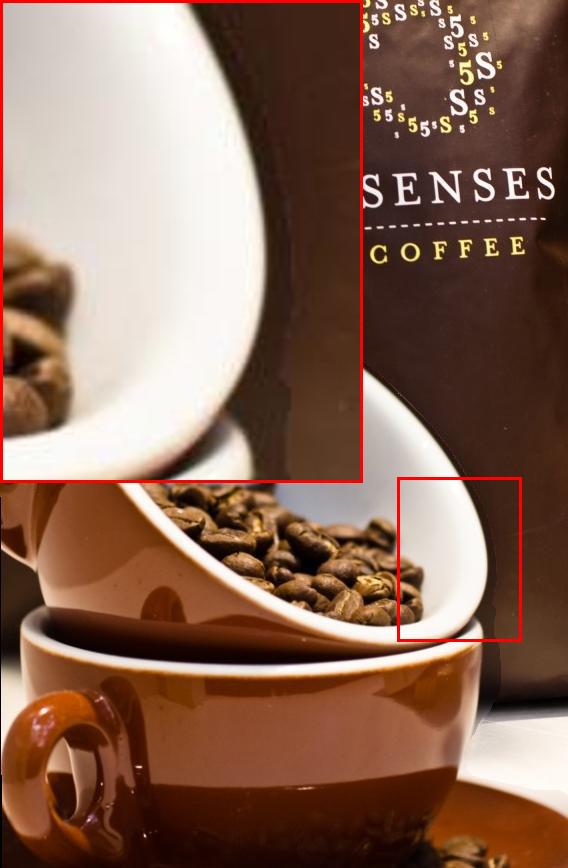}}
  \subfigure[Ground truth]{
  \includegraphics[scale = 0.11]{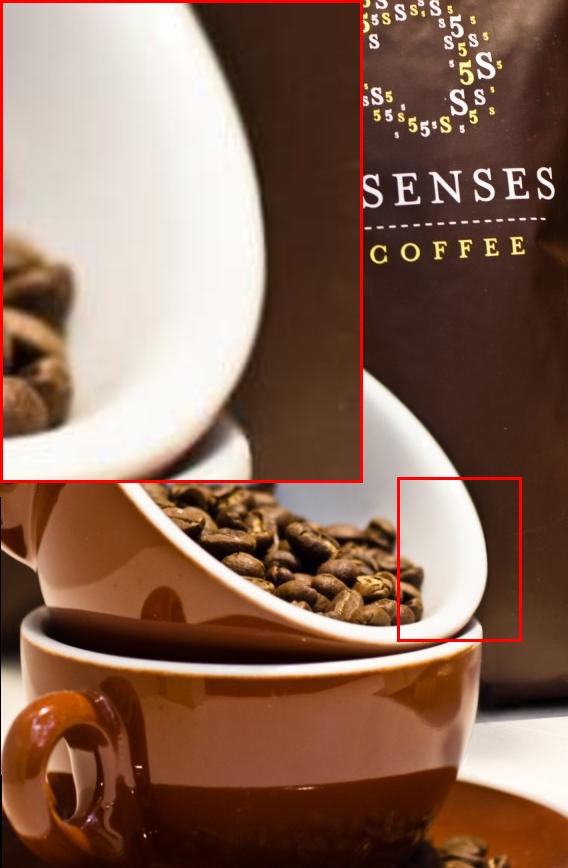}}
  \caption{\footnotesize The fusion results on Lytro (20th) and MFFW (11th) with detail magnified of all algorithms. 
  The FuseGAN\_Ib exhibits the fusion result
  by FuseGAN with ConvCRF which feed with $I_{B}$ as ground truth.}
  \label{Fig_visual_compare_sota}
\end{figure*}

More attention could be paid to the details on the basis that the 
overall fusion results are good. So based on the 20th image in Lytro and 
the 11th in MFFW, the fusion results with detailed magnified of all methods,
as shown in Fig.\ref{Fig_visual_compare_sota}, are 
compared visually to show the superior fusion performance of MFIF-GAN, 
especially in the region around the FDB.

Furthermore, by comparing the Fig.\ref{Lytro20_FuseGAN} and \ref{Lytro20_FuseGAN_Ib}, 
\ref{MFFW11_FuseGAN} and \ref{MFFW11_FuseGAN_Ib} respectively,
it can be seen that if $I_{A}$, as the input of ConvCRFs in the FuseGAN,
is replaced with $I_{B}$ suffers from the DSE, the edge of the foreground
will be much clearer.

\subsubsection{Solution to mitigating the DSE}

As we discussed in the drawbacks of FuseGAN, when the all-in-focus image do not exist, 
the $I_{A}$ has to be used as ground truth for the ConvCRFs.
However, the DSE vanishes when it comes to $I_{A}$,
which means the FDB in $I_{A}$ is definitely clear.
Thus the outputs of FuseGAN are focus maps which represent the real foreground
objects with sharp edges.
According to the $\alpha$-matte model theory and the experience in 
daily observation, using these focus maps could result in two completely different 
situations in the procedure of extracting clear regions from source images:

When $I_{A}$ is processed, the extracted foreground region is ideal.
In contrast, when it comes to $I_{B}$, because of the existence of the DSE, 
there must be a part of the foreground information diffuses into the background.
So the diffussion laying outside the foreground objects remain
in the clear background (as shown obviously in the Fig.\ref{Lytro20_FuseGAN} 
and \ref{MFFW11_FuseGAN}).
Consequently, the fuzzy FDB appears in the fusion images, which is the diffusion of the
foreground in $I_{B}$ essentially.
That also partially explains why the undesirable results can be alleviated 
to some extend in FuseGAN when $I_{A}$ is substituted with $I_{B}$,
as shown in Fig.\ref{Lytro20_FuseGAN_Ib} and \ref{MFFW11_FuseGAN_Ib}.

Actually, a part of the information about clear backgrounds
is indeed missed as it is covered by the diffusion of foregrounds
in $I_{B}$.  Moreover, these regions are irregular,
as the width is affected by the shape of foregrounds and the distance 
from different positions to the sensor. So it is extremely complex and 
nontrivial to ideally handle FDB regions and eliminate the DSE.

One of the solutions is to generate focus maps in which the foregrounds
are mildly larger than real objects. Using these focus maps can remain 
the backgrounds information around foregrounds in $I_{A}$.
Actually, focus maps obtained by MFIF-GAN is exactly what we expected.

\begin{figure}[H]
  \centering
  \subfigure[]{
    \includegraphics[scale = 0.166]{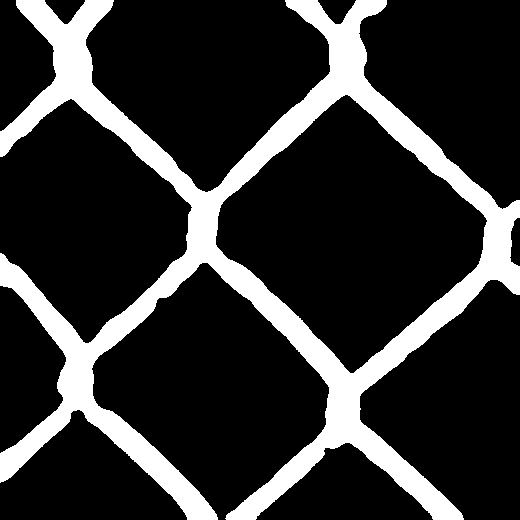}}
  \subfigure[]{
  \includegraphics[scale = 0.166]{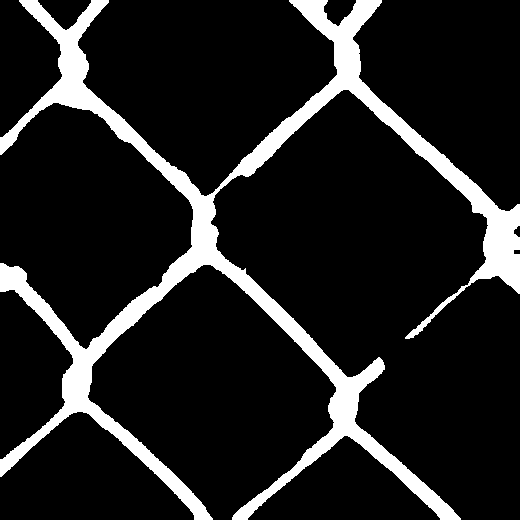}}
  \subfigure[]{
  \includegraphics[scale = 0.166]{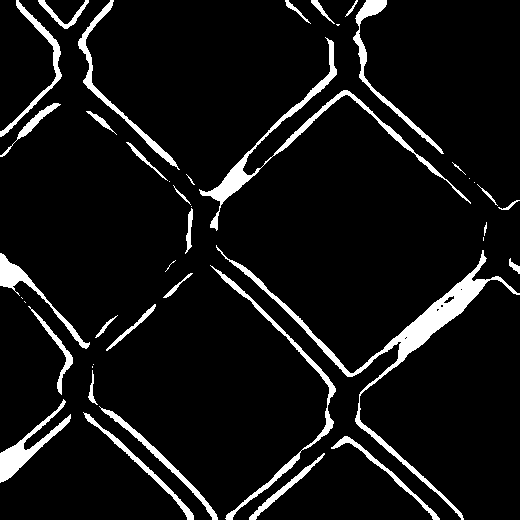}\label{diff_focus_map}}
  \subfigure[]{
  \includegraphics[scale = 0.1224]{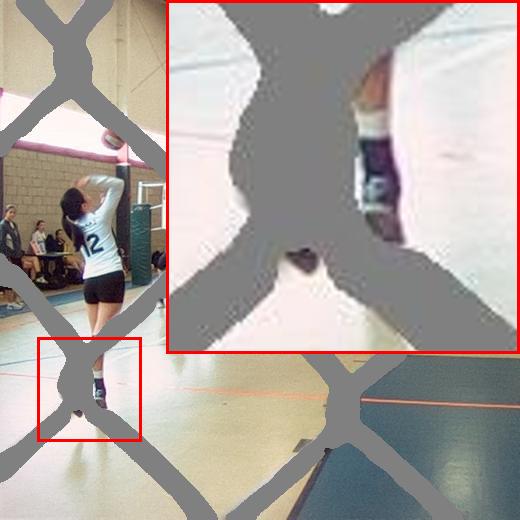}\label{MFIF-GAN_background}}
  \subfigure[]{
  \includegraphics[scale = 0.1224]{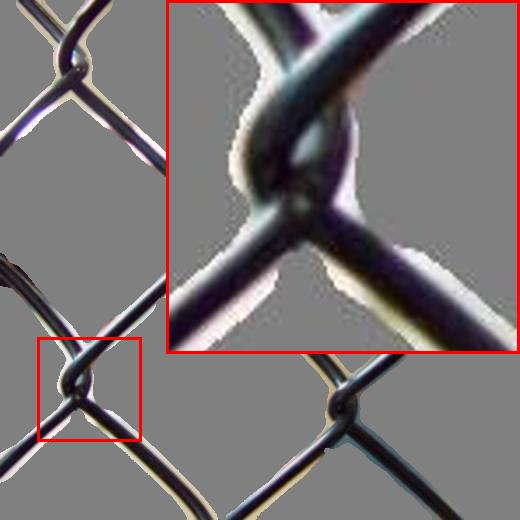}\label{MFIF-GAN_foreground}}
  \subfigure[]{
  \includegraphics[scale = 0.1224]{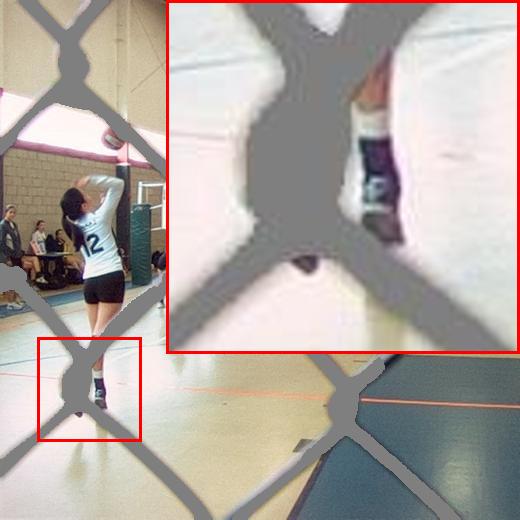}\label{FuseGAN_background}}
  \subfigure[]{
  \includegraphics[scale = 0.1224]{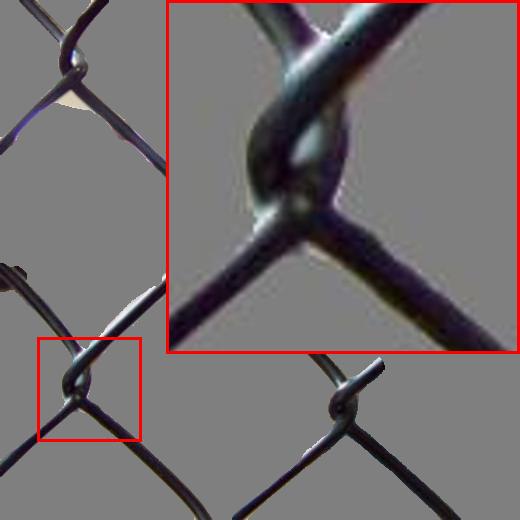}\label{FuseGAN_foreground}}
  \subfigure[]{
    \includegraphics[scale = 0.257]{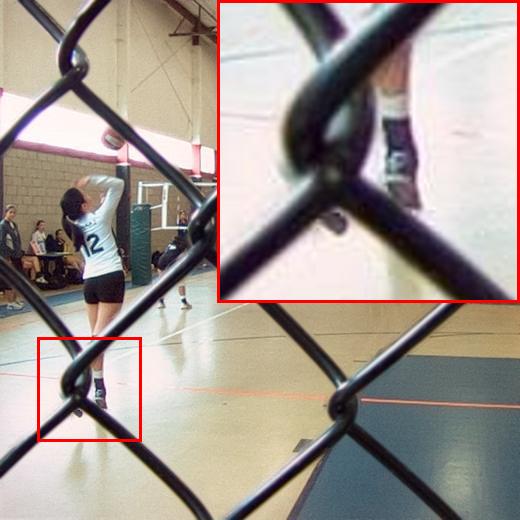}}
  \subfigure[]{
    \includegraphics[scale = 0.257]{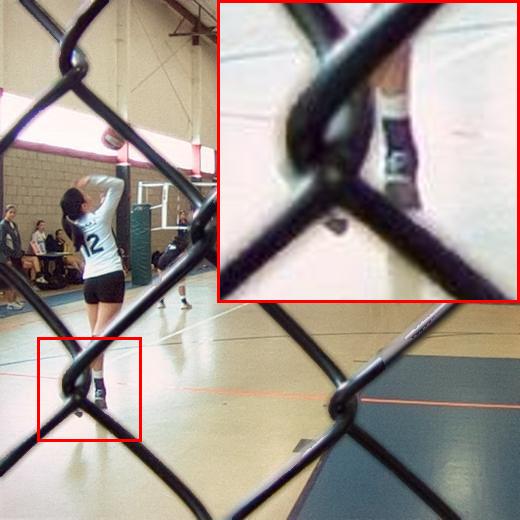}}
  \caption{\footnotesize (a) and (b) exhibits the $\hat{F}_{final}$ 
generated by MFIF-GAN and FuseGAN respectively; (c) is the 
difference between two focus maps; (d) and (e) shows the background and foreground 
extracted respectively by $\hat{F}_{final}$ in (a); (f) and (g) are the 
counterparts processed by $\hat{F}_{final}$ in (b);
(h) and (i) are fusion results of MFIF-GAN and FuseGAN with detail magnified.}
\end{figure}

To illustrate this statement, we make the difference between 
two sets of focus maps generated by baseline FuseGAN and MFIF-GAN respectively. 
As shown in Fig.\ref{diff_focus_map}, the appearance of white edges indicates 
this statement obviously.

To show the improvement brought by this charactoristic more clearly and intuitively, 
we used these two focus maps respectively to extract the foreground and background
of source images. As shown in Fig.\ref{MFIF-GAN_background} and \ref{MFIF-GAN_foreground}, 
the background extracted by the focus map generated by MFIF-GAN partially eliminates 
the edge diffsuion, that is DSE. Meanwhile the extracted foreground contains 
part of the blurred background which can smooth the FDB.
In contrast, as shown clearly in Fig.\ref{FuseGAN_background} 
and \ref{FuseGAN_foreground}, even though the
foreground extracted by the focus map generated by FuseGAN 
seems to be ideal, the extracted background in Fig.\ref{FuseGAN_background} 
retains foreground diffusion around the edge.

\subsubsection{Diffusion and Contraction Experiments}

\begin{figure}[H]
  \centering
  \subfigure[-4 pixels]{
    \includegraphics[scale = 0.1]{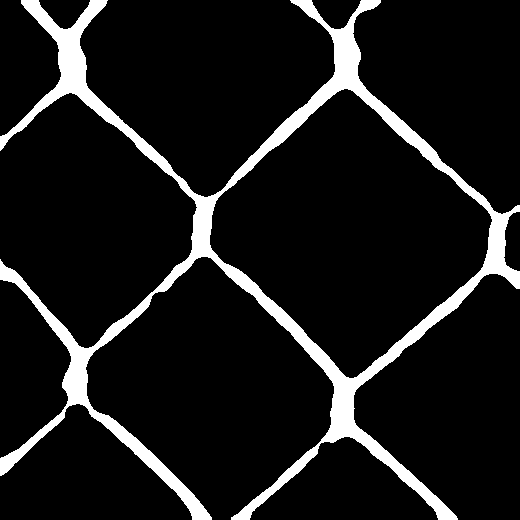}\label{-4 pixels}}
  \subfigure[-2 pixels]{
    \includegraphics[scale = 0.1]{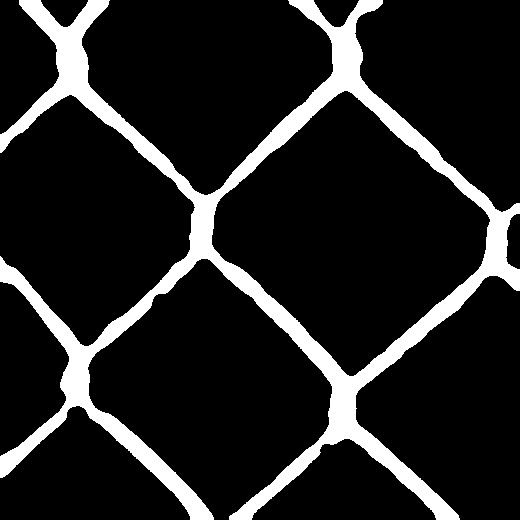}\label{-2 pixels}}
  \subfigure[origin]{
    \includegraphics[scale = 0.1]{focus_map_pp5_MFIF-GAN.png}\label{origin}}
  \subfigure[2 pixels]{
    \includegraphics[scale = 0.1]{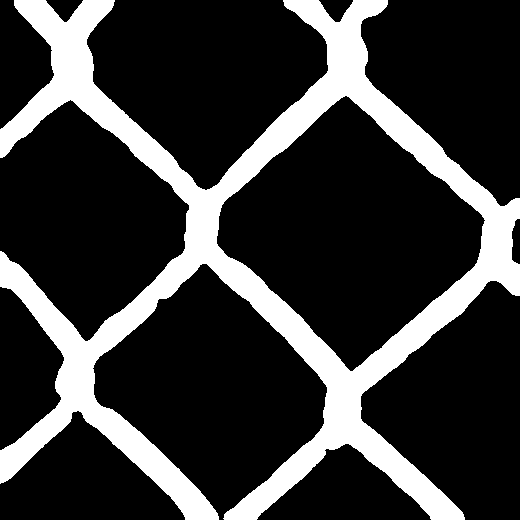}\label{2 pixels}}
  \subfigure[4 pixels]{
    \includegraphics[scale = 0.1]{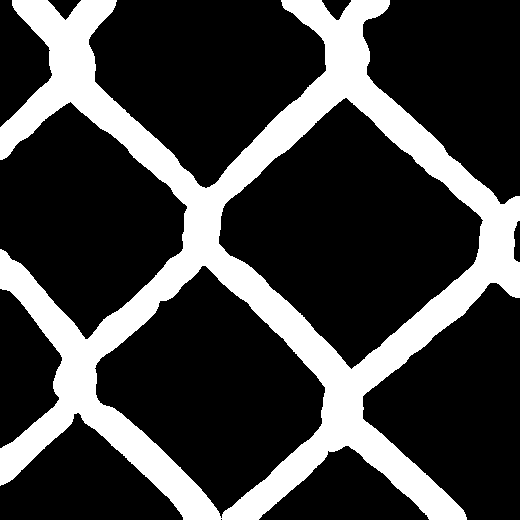}\label{4 pixels}}
  \subfigure[]{
    \includegraphics[scale = 0.66]{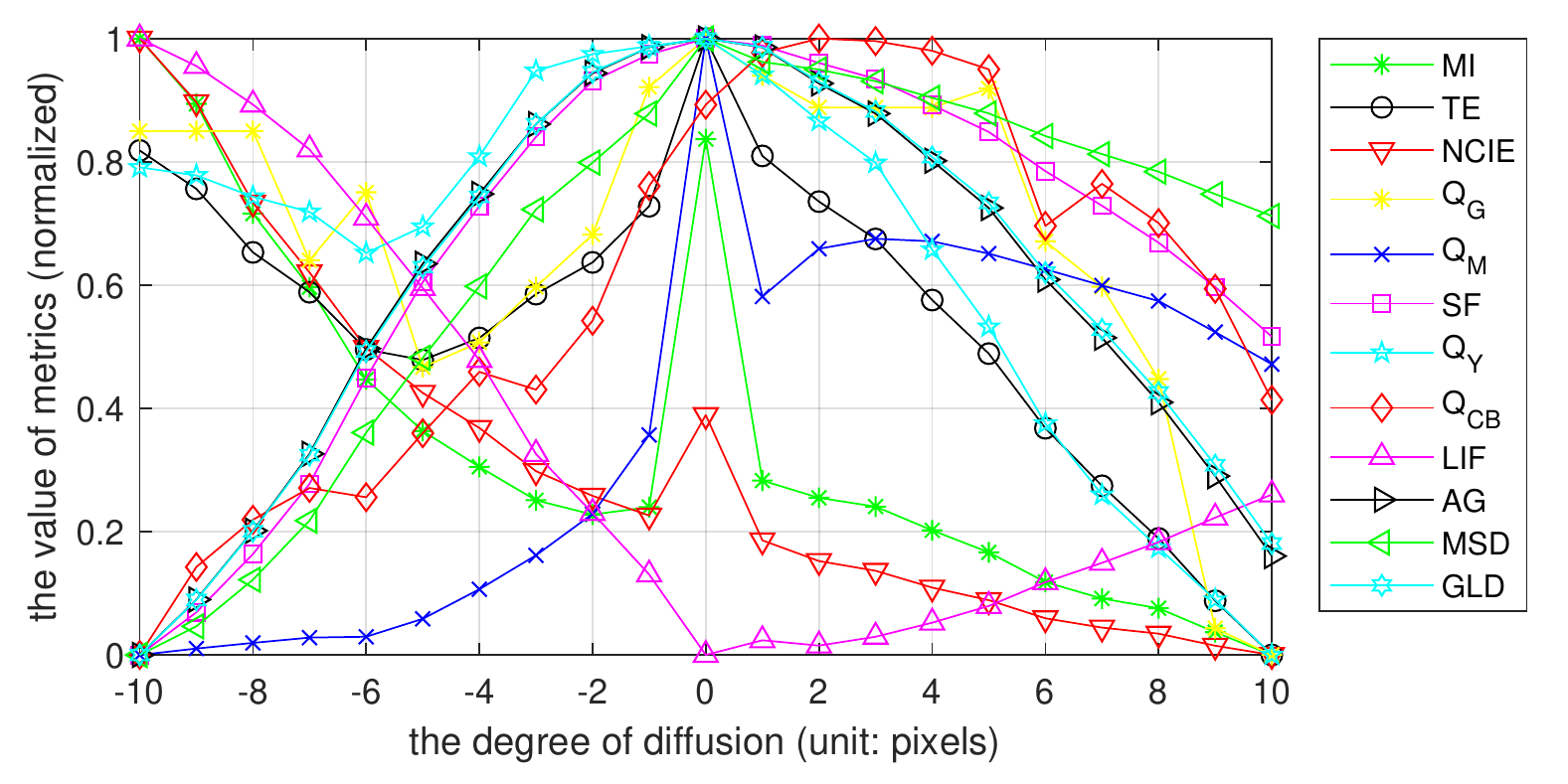}\label{the_influence_of_diffusion}}
  \caption{\footnotesize (a)-(e) are focus maps with different degree of edge diffusion
  or contraction in which (c) is the original focus map generated by our method. And
  (f) is the average performance on Lytro using these focus maps. All
  values of each metrics are normalized to [0-1].}
  \label{Diffusion and Contraction}
\end{figure}

The scale of the foreground in focus maps is crucial
when we employ proposed solution to attenuate the DSE.
In order to explore the rational size of the foreground regions and confirm 
the effectiveness of MFIF-GAN, we design a diffusion and contraction 
module to enlarge or shrink the foreground area in focus maps generated
by our method at the pixel level. As shown in Fig.\ref{Diffusion and Contraction}
(a)-(e), the negative value of the degree of diffusion
indicates foreground regions are contracted by certain pixels.
According to equation (\ref{fuse_equation}), a series of focus maps
with differnet degree of edge diffusion or contraction are used as guidance
to fuse source images.

As reported in Fig.\ref{the_influence_of_diffusion},
the results assessed by $MI$ and $NCIE$ indicate that apart from using original
focus maps, the performances decrease with the expansion of the
foreground regions. And the line of $Q_{CB}$ shows that the foreground regions
which are two pixels larger than the original ones could bring best performance.
Except for these three metrics, experimental results generally
illustrate that using the focus maps without any operations achieves
the largest values. That demonstrates MFIF-GAN can produce focus maps 
which have accurate size of foreground regions and attenuate the DSE exactly 
at the pixel level.

\subsubsection{Execution Time}
This section is about the comparison of computational efficiency.
Tab.\ref{time_consumption} lists the mean execution time of each method 
on testing datasets\footnote{Because the fusion results of MMF-net on Lytro
are used directly, this method is not involved in the comparison.}.
The experiments are carried out on a computer 
with Intel Core i7-10700K CPU @ 3.8GHz and RTX 2080ti GPU.
The results indicate that compared with SOTAs, the fusion efficiency of our
algorithm is the highest.

\begin{table*}[htbp]
  \small
  \centering
  \caption{\footnotesize Average used time of all methods for pre-pair images 
  fusion (unit: seconds)}
  \resizebox{\textwidth}{!}{
  \begin{tabular}{ccccccccc}
    \toprule
     ~ & CNN & MWGF & Quadtree  & DSIFT & CSR & NSCT& FuseGAN & MFIF-GAN\\
     \midrule
    Lytro &25.6188 & 1.9677 &0.5416  &0.9095 &120.9291 & 1.7989 &0.4976 &\textbf{0.2229}\\
  
    MFFW &28.6874 & 2.2521 &0.4922  &1.6303 & 139.3664 & 2.2881 &0.5019 &\textbf{0.2236}\\

    grayscale &18.0621 & 0.4808 &0.3029 &0.7909 & 28.9021 & 0.4929 & 0.2883 &\textbf{0.1344}\\
    \bottomrule
  \end{tabular}}
  \label{time_consumption}
\end{table*}

FuseGAN needs extra post-processing, so we record the time with respect to
the generation of initial focus maps, post-processing
and final fusion.
The average time used are
(Lytro) 0.2135s, 0.2355s, 0.0486s;
(MFFW) 0.2156s, 0.2393s, 0.047s;
(grayscale) 0.078s, 0.1969s, 0.0133s respectively.

\subsection{Ablation Experiments}

\begin{figure}[H]
  \centering
  \includegraphics[scale = 0.3]{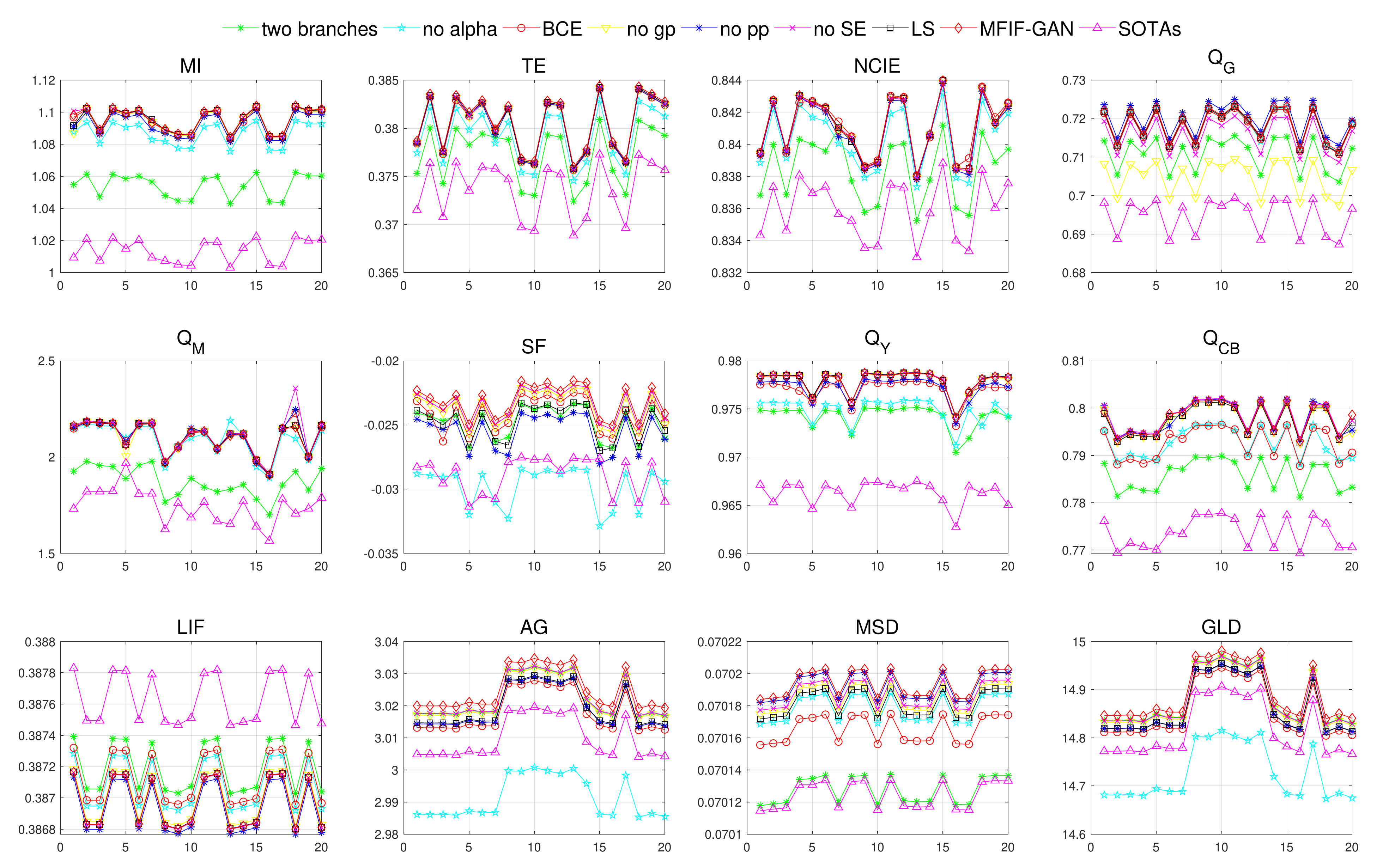}
  \caption{\footnotesize Score lines of ablation experiments based on Lytro 
  with regard to each metric. The lines named ``SOTAs'' are average scores in terms
  of all SOTAs on every image.}
  \label{Ablation_result}
\end{figure}

The innovations of our MFIF-GAN include a new encoder with six parallel branches as well as 
attention modules, the well-designed loss function with gradient regularization and
$\ell_1$-norm based reconstruction, and a simple but effective post-processing strategy.
In addition, the network is trained on an $\alpha$-matte dataset.
To validate the role of each element in our work, a series of ablation experiments 
are conducted here. The results are shown in Fig.\ref{Ablation_result}.

(1) In order to verify the capability of our MFIF-GAN to extract features
of color images, the structure of encoder in generator are redesigned 
with two parallel sub-networks which input images with single channel.
(2) To show the effectiveness of the $\alpha$-matte model,
our network is also trained on the conventional MFIF dataset generated in section
\ref{dataset_generation}. From score lines named ``two branches'' and ``no alpha''
as shown in every sub-figure, we can see that apart from the SOTAs, 
the network with two branches and MFIF-GAN 
trained on conventional MFIF dataset are generally
inferior to all other counterparts in ablation experiments,
which indicates that the structure of the network and training data play significant
roles and more implicit features could be extracted in color images suffering from
the DSE.

(3) The $\ell_1$-norm based reconstruction loss $\mathcal{L}_{rec}$ 
(equation (\ref{reconstruction})) is 
substituted by the BCE used in baseline work to testify the validity of this loss.
Compared with other sub-figures, the worse performances of line ``BCE'' 
on edge detail assessment metrics 
($LIF$, $AG$, $MSD$ and $GLD$) indicate that the $\ell_1$-norm based 
loss $\mathcal{L}_{rec}$ could force the generator to pay more attention 
to the information around the FDB and enhance the edge details.

(4) The gradient penalty $\mathcal{L}_{gp}$ (equation (\ref{gradient_penalty})),
and post-processing procedure 
are removed respectively in MFIF-GAN. 
Also, the SE-ResNet block is replaced by general ResNet
to prove the improvement brought by this attention mechanism.
In addition, to show the role of adversarial loss $\mathcal{L}_{adv}$ (equations
(\ref{adv_D}) and (\ref{adv_G}))
rised in \cite{improved_WGAN},
they are replaced by the least squares loss \cite{LSGAN} used in the FuseGAN.
The lines named ``no gp'', ``no pp'', ``no SE'' and ``LS'' show that each of these factors 
improves the performance of fusion results to some extent, 
but generally none of them is more important than others.

Last but not least, as shown in score lines ``SOTAs'', it is worthy to note that
without any one of factors, our algorithm still has a big advantage over other 
SOTAs generally, which indicates that proposed MFIF-GAN with a new architecture 
and well-designed loss function has strong robustness for gradient regularization, 
post-processing, attention mechanism and training dataset.


\section{Conclusions}
\label{Conclusions}
In this paper, we propose a generative adversarial network termed MFIF-GAN 
for the MFIF task and put forward a new solution to attenuate the DSE which is
rampant in this field. The motivation of our work is to guarantee the 
foreground regions in generated focus maps are mildly larger than the 
corresponding objects, which can simulate the DSE and further exactly 
alleviate this annoying effect.
The innovation is that attention machanism is exploited 
in the network which has a new architecture with six branches to extract features. 
And the $\ell_1$-norm reconstruction loss and gradient penalty is 
creatively added to the optimization function to enhance the edge details and
improve the quality of the outputs.
Moreover, the SRR algorithm for post-processing is used to refine the initial 
focus maps in a computational effective way.
Last but not least, based on a synthetic $\alpha$-matte training dataset, 
this novel end-to-end color multi-focus image fusion algorithm can fuse more 
realistic images especially around the FDB.

As a new fusion algorithm, experiments demonstrate that
our MFIF-GAN is superior to other representative SOTA methods
on visual perception, quantitative analysis and efficiency.
The edge diffusion and contraction module proves that
following the proposed solution, MFIF-GAN can generate accurate focus maps
and alleviate the DSE at the pixel level, which can bring more
satisfactory pretreatment to other computer vision tasks.

\section*{Acknowledgements}

The research is supported by the National Key Research and Development Program
of China under grant 2018AAA0102201, the National Natural Science Foundation
of China under grant 61877049, 61976174 and 11671317.


\bibliography{citationlist}

\end{document}